\documentclass[letterpaper]{article} % DO NOT CHANGE THIS
\usepackage{aaai23}  % DO NOT CHANGE THIS
\usepackage{times}  % DO NOT CHANGE THIS
\usepackage{helvet}  % DO NOT CHANGE THIS
\usepackage{courier}  % DO NOT CHANGE THIS
\usepackage[hyphens]{url}  % DO NOT CHANGE THIS
\usepackage{graphicx} % DO NOT CHANGE THIS
\usepackage{natbib}  % DO NOT CHANGE THIS AND DO NOT ADD ANY OPTIONS TO IT
\usepackage{caption} % DO NOT CHANGE THIS AND DO NOT ADD ANY OPTIONS TO IT
\usepackage{algorithm}
\usepackage{algorithmic}
\usepackage{multirow}

\usepackage{amssymb}
\usepackage{amsmath}
\usepackage{enumitem}

\usepackage{newfloat}
\usepackage{listings}
\DeclareCaptionStyle{ruled}{labelfont=normalfont,labelsep=colon,strut=off} % DO NOT CHANGE THIS
\floatstyle{ruled}
\newfloat{listing}{tb}{lst}{}
\floatname{listing}{Listing}
\nocopyright
\title{MIGA: A Unified Multi-task Generation Framework for Conversational Text-to-SQL}

\author{
    Yingwen Fu\textsuperscript{\rm 1\rm 2}\equalcontrib\thanks{This work was conducted when Yingwen Fu was interning at NetEase Games AI Lab.}, Wenjie Ou\textsuperscript{\rm 2}\equalcontrib \thanks{Corresponding author.}, Zhou Yu\textsuperscript{\rm 3}, and Yue Lin\textsuperscript{\rm 2}
}
\affiliations {
    \textsuperscript{\rm 1} Guangdong University of Foreign Studies, Guangzhou, China\\ 
    \textsuperscript{\rm 2} NetEase Games AI Lab, Guangzhou, China\\
    \textsuperscript{\rm 3} Columbia University\\
    20201010002@gdufs.edu.cn, zy2461@columbia.edu \\
    \{ouwenjie, gzlinyue\}@corp.netease.com
}
\usepackage{bibentry}
\begin{document}

\maketitle

\begin{abstract}
Conversational text-to-SQL is designed to translate multi-turn natural language questions into their corresponding SQL queries. Most state-of-the-art conversational text-to-SQL methods are incompatible with generative pre-trained language models (PLMs), such as T5. In this paper, we present a two-stage unified \textbf{M}ult\textbf{I}-task \textbf{G}eneration fr\textbf{A}mework (MIGA) that leverages PLMs' ability to tackle conversational text-to-SQL. In the pre-training stage, MIGA first decomposes the main task into several related sub-tasks and then unifies them into the same sequence-to-sequence (Seq2Seq) paradigm with task-specific natural language prompts to boost the main task from multi-task training. Later in the fine-tuning stage, we propose four SQL perturbations to alleviate the error propagation problem. MIGA tends to achieve state-of-the-art performance on two benchmarks (SparC and CoSQL). We also provide extensive analyses and discussions to shed light on some new perspectives for conversational text-to-SQL.
\end{abstract}

\section{Introduction}
\begin{figure}
  \centering
  \includegraphics[width=0.4\textwidth]{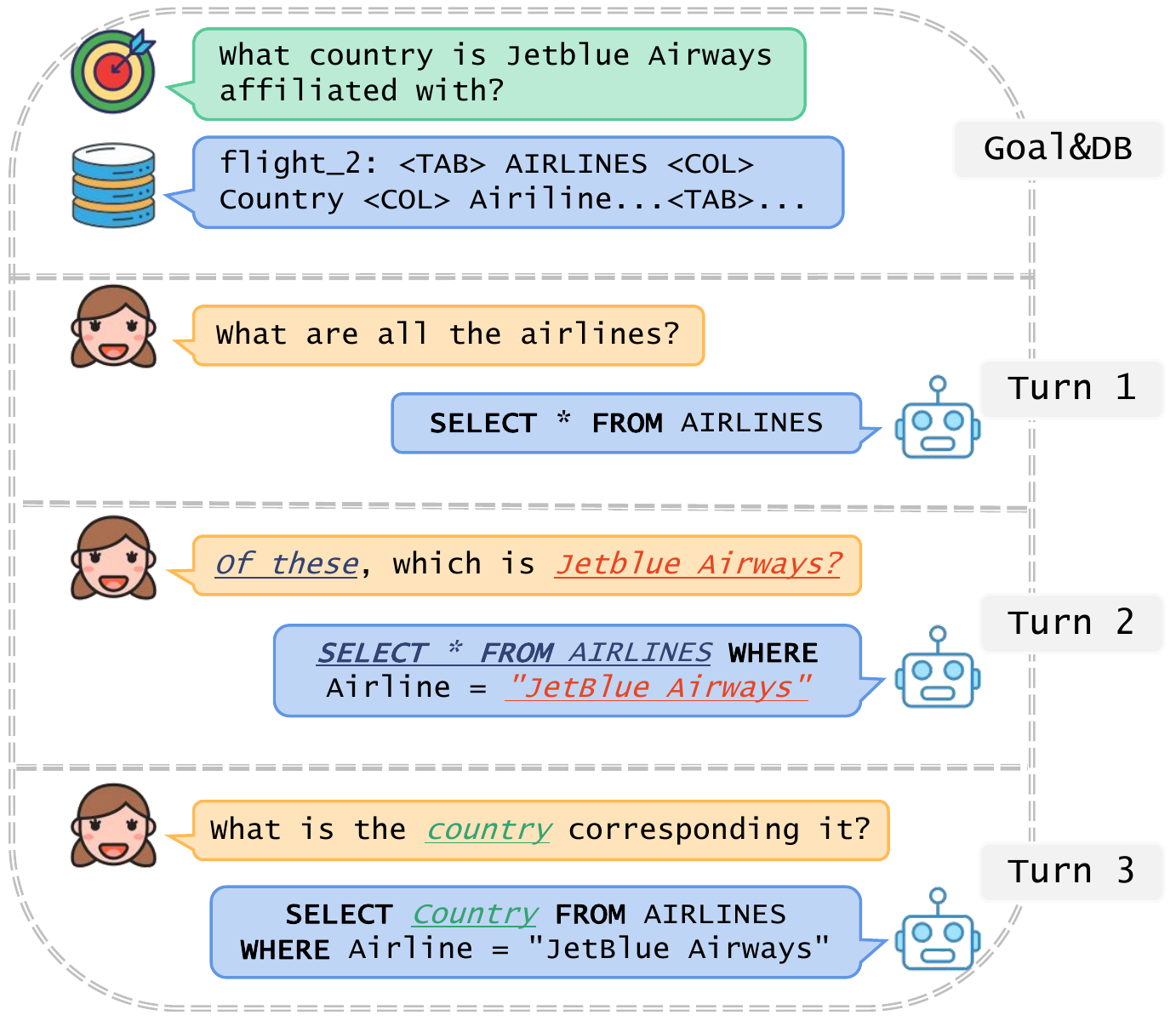} 
  \caption{An example of conversational text-to-SQL.} 
  \label{fig:Example} 
\end{figure}

As an important branch of semantic parsing, the text-to-SQL task \cite{DBLP:journals/corr/abs-1709-00103, DBLP:journals/corr/abs-1711-04436} aims to automatically generate SQL queries according to natural language questions. It could break through the barrier for non-expert users to write SQL queries and interact with database systems. While in practical scenarios, users tend to obtain target information from the system step-by-step through conversations. To meet this demand, conversational text-to-SQL has recently drawn considerable attention. As shown in Figure \ref{fig:Example}, it extends text-to-SQL into multi-turn settings by generating interactive queries based on conversations.

Recently, generative pre-trained language models (PLMs) represented by T5 \cite{DBLP:journals/jmlr/RaffelSRLNMZLL20} achieve competitive performance on multiple sequence-to-sequence (Seq2Seq) tasks. However, as a typical Seq2Seq task, most advanced methods \cite{cai2020igsql, hui2021dynamic, DBLP:conf/acl/ChenCLCMWY21, DBLP:conf/iclr/0009ZPMA21, DBLP:conf/acl/ZhengWDWL22, DBLP:journals/corr/abs-2112-08735, DBLP:journals/corr/abs-2205-07686} for conversational text-to-SQL are based on tree-based decoders \cite{DBLP:conf/acl/YinN17, DBLP:conf/emnlp/ScholakSB21}. While effective, these methods require custom transition grammars which need additional overhead when being extended to new SQL structures. Additionally, the custom decoder structures make them incompatible with the generative PLMs. On the contrary, some recent attempts such as UNIFIEDSKG \cite{DBLP:journals/corr/abs-2201-05966}, PICARD \cite{DBLP:conf/emnlp/ScholakSB21}, and RASAT \cite{DBLP:journals/corr/abs-2205-06983} have witnessed the potential of generative PLMs for this task. Another line of research on language pre-training \cite{DBLP:journals/corr/abs-1811-01088, DBLP:conf/emnlp/AghajanyanGSCZG21, DBLP:conf/acl/SuSMG0LZ22, DBLP:journals/corr/abs-2204-06923} demonstrates that pre-training with intermediate-labeled data of related tasks can help improve the fine-tuned models. This suggests that we could boost the model for conversational text-to-SQL by pre-training on its related tasks. 

\begin{figure*}
  \centering
  \includegraphics[width=1\textwidth]{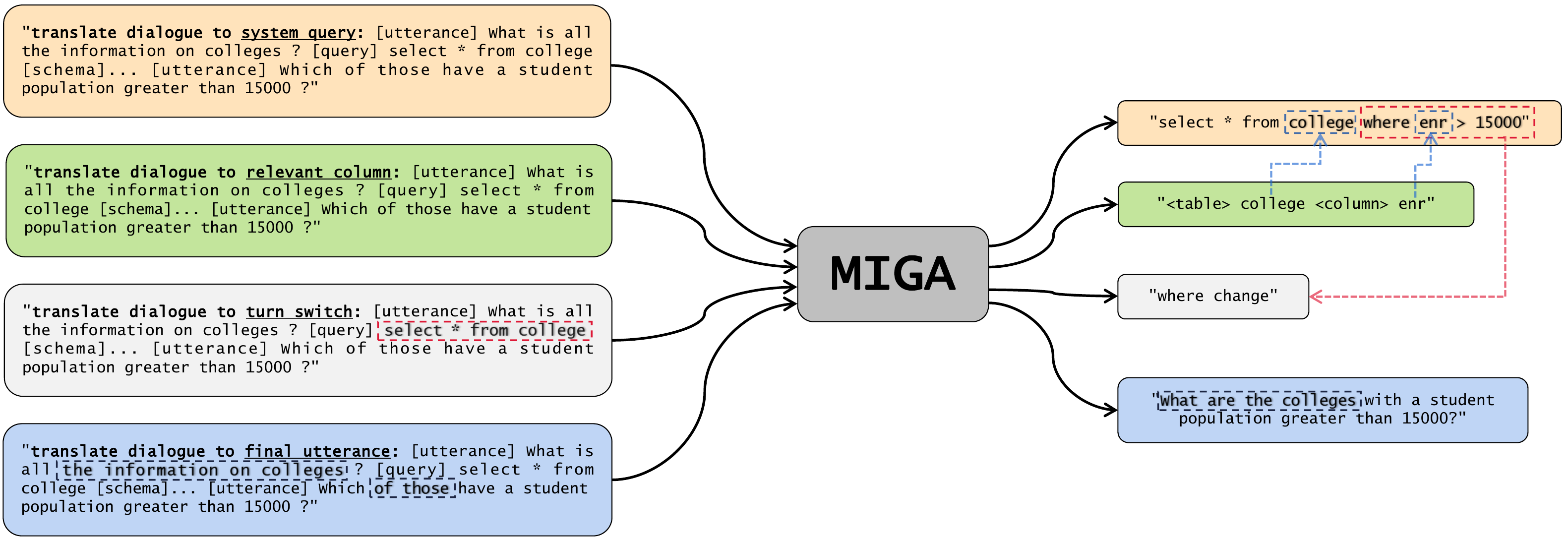} 
  \caption{Overview of the proposed framework. In the multi-task pre-training stage, we train the model with four tasks: SQL Generation (SG, orange box), Related Schema Prediction (RSP, green box), Turn Switch Prediction (TWP, gray box), and Final Utterance Prediction (FUP, blue box). We unify them into the same Seq2Seq paradigm and jointly train them with task-specific prompts.{\tt "[utterance]"},{\tt "[query]"}, and {\tt "[schema]"} respectively indicate the user utterance, SQL query, and the schema of the target database. } 
  \label{fig:1} 
\end{figure*}

In this paper, inspired by in-context learning \cite{DBLP:conf/nips/BrownMRSKDNSSAA20}, we propose a two-stage unified \textbf{M}ult\textbf{I}-task \textbf{G}eneration fr\textbf{A}mework (MIGA) for conversational text-to-SQL based on the pre-trained T5 models. MIGA consists of multi-task pre-training and single-task fine-tuning stages. In the pre-training stage (Stage 1), we first decompose the main task conversational text-to-SQL (abbreviated as SQL generation (SG) henceforth) into three sub-tasks and then unify them in a Seq2Seq paradigm by inserting task-specific natural language prompts into the input text to indicate each task. The model is jointly trained in the main task as well as three sub-tasks in this stage.
\begin{itemize}
    \item \textbf{Related Schema Prediction (RSP)}. A common error in the generated SQL is generating wrong tables and columns. Inspired by SCoRe \cite{DBLP:conf/iclr/0009ZPMA21}, RSP is defined as generating the tables and columns that appeared in the current SQL query. This sub-task only needs to consider the to-be-appeared columns and tables without taking the generated SQL structure into account, which can effectively alleviate the above issue. As shown in Figure \ref{fig:1}, the RSP output for the current turn indicates the appeared table {\tt "college"} and column {\tt "enr"} in the SQL query {\tt "select * from college where enr > 15000"}. 
    \item \textbf{Turn Switch Prediction (TWP)}. Followed by \cite{DBLP:journals/corr/abs-2112-08735}, TWP is introduced to enhance the model by modeling the conversation flow between the adjacent SQL query pair. This task is to predict what modification is made to the current SQL compared with its last SQL. In Figure \ref{fig:1}, the current TWP output indicates that compared with the last SQL, the WHERE sub-clause is modified as {\tt "where enr $\mathcal{>}$ 15000"}. Thus we simplify this modification as {\tt "where change"}. 
    \item \textbf{Final Utterance Prediction (FUP)}. As stated in \cite{DBLP:journals/corr/abs-2205-07686}, CQR can effectively improve the model by alleviating the contextual dependency. In this paper, similar to CQR, we propose FUP to generate the final utterance for the whole interaction. The FUP output in Figure \ref{fig:1} summarizes the information of the two-turn conversation by replacing {\tt "which of those"} with {\tt "what are the colleges"}.
\end{itemize}

Such a unified framework provides a general and flexible paradigm that can handle any tasks whose input and output can be recast as a sequence of tokens. It serves some advantages: (1) referring to the steps humans write SQL, the conversational text-to-SQL task is decomposed into multiple sub-tasks, allowing the main task to learn from them; (2) the training samples are constructed as the format of T5 which can maximize the potential of the pre-trained T5 model for the target task; and (3) the unified framework allows plug-and-play of multiple auxiliary tasks. When applying to a specific task, we fine-tune the pre-trained model above (Stage 2) using the same training objective in the task-specific labeled data. When predicting the current-turn SQL, the previous generated SQLs are concatenated in the current model input, which may cause error propagation problem. To alleviate this, we propose to perturb the previous-turn ground truth SQLs according to four common generation errors, aiming to reduce the dependency of previous SQLs.

We evaluate MIGA on two conversational text-to-SQL benchmarks. Comparisons against previous advanced approaches show that MIGA tends to achieve state-of-the-art (SOTA) in two datasets based on interaction match (IM) accuracy. In summary, our contributions are:
\begin{enumerate}
    \item We propose a novel framework, namely MIGA, that decomposes conversational text-to-SQL into three sub-tasks and unify them into a Seq2Seq paradigm. This can effectively facilitate the main task learning. The auxiliary tasks are trained in a plug-and-play manner, which makes the framework easily accommodate more auxiliary tasks.
    \item We propose four SQL perturbations when fine-tuning the model, which can effectively alleviate the error propagation problem.
    \item Experimental results on two benchmark datasets show that MIGA tends to achieve SOTA performance on IM accuracy. Extensive analyses provide some new insights into the proposed framework and some takeaways for future conversational text-to-SQL studies.   
\end{enumerate}

\section{Related Work}
\subsection{Text-to-SQL Task}
As a representative Seq2Seq task, context-independent text-to-SQL represented by Spider \cite{DBLP:conf/emnlp/YuZYYWLMLYRZR18} benchmark has witnessed multiple meaningful advances in encoder  \cite{DBLP:conf/acl/WangSLPR20, DBLP:conf/nips/CaiYXH21, DBLP:conf/acl/CaoC0ZZ020}, decoder \cite{DBLP:conf/acl/YinN17, DBLP:conf/emnlp/ScholakSB21}, and language pre-training \cite{DBLP:conf/naacl/DengAMPSR21, DBLP:conf/iclr/0009WLWTYRSX21, DBLP:conf/iclr/0009ZPMA21}. With the widespread popularity of dialogue research, conversational text-to-SQL benchmarks SparC \cite{DBLP:conf/acl/YuZYTLLELPCJDPS19} and CoSQL \cite{DBLP:conf/emnlp/YuZELXPLTSLJYSC19} have drawn much attention. In particular, EditSQL \cite{DBLP:conf/emnlp/ZhangYESXLSXSR19} and IST-SQL \cite{DBLP:conf/aaai/WangLZ021} propose to leverage previous generated SQL to improve the SQL quality in the current turn. DELTA \cite{DBLP:conf/acl/ChenCLCMWY21} and CQR-SQL \cite{DBLP:journals/corr/abs-2205-07686} focus on the application of conversational question reformulation (CQR) \cite{DBLP:conf/emnlp/PanBWZL19} and context-independent text-to-SQL to alleviate the difficulty of multi-turn SQL generation. RAT-SQL-TC \cite{DBLP:journals/corr/abs-2112-08735} and HIE-SQL \cite{DBLP:conf/acl/ZhengWDWL22} aim to utilize the interaction information between the current and previous text as well as the SQL query to enhance the model performance respectively by two auxiliary objectives and a bimodal pre-trained model of natural language and SQL. 

Recently, generic Seq2Seq pre-trained language models such as T5\cite{DBLP:journals/jmlr/RaffelSRLNMZLL20} have unveiled their potential for multiple generation tasks. UNIFIEDSKG \cite{DBLP:journals/corr/abs-2201-05966} fine-tunes a pre-trained T5-3B model with text-to-SQL data that could yield results competitive to the then-state-of-the-art. Based on the T5 architecture, RASAT \cite{DBLP:journals/corr/abs-2205-06983} uses an augmented relation-aware self-attention to enhance the interaction between the database schema and questions. Different from the methods above, this paper concentrates on improving the conversational text-to-SQL performance without additional annotated training data or additional model structure.  

\subsection{Pre-training on Supplementary Tasks}
When further training is applied to tasks containing intermediate-labeled data \cite{DBLP:journals/corr/abs-1811-01088, DBLP:conf/emnlp/AghajanyanGSCZG21, DBLP:conf/acl/SuSMG0LZ22, DBLP:journals/corr/abs-2204-06923}, the fine-tuned model has been shown to perform much better on a variety of tasks, including the GLUE benchmark \cite{DBLP:conf/iclr/WangSMHLB19} and conversational recommender systems \cite{DBLP:conf/wsdm/Lei0MWHKC20, DBLP:conf/nips/LiKSMCP18}. Specifically, PPTOD \cite{DBLP:conf/acl/SuSMG0LZ22} proposes to enhance the task-oriented dialogue system by supplementarily training on three sub-tasks: end-to-end dialogue modeling, dialogue state tracking, and intent classification. Inspired by PPTOD, we study a similar supplementary training setup with three well-designed sub-tasks for conversational text-to-SQL.

\section{Methodology}
Algorithm \ref{alg:algorithm} gives the overall training process of the proposed MIGA framework in this paper, which consists of multi-task pre-training and single-task fine-tuning stages.
% The proposed MIGA framework consists of multi-task pre-training and single-task fine-tuning stages. The training algorithm is summarized in Appendix A.

\begin{algorithm}[H]
\caption{Training Process of MIGA}
\label{alg:algorithm}
\textbf{Input}: Training set $\mathcal{D}_1$=$\{\mathcal{D}_{SG}, \mathcal{D}_{RSP},\mathcal{D}_{TWP}, \mathcal{D}_{FUP}\}$\\=$\{(Z_T, X, y)_i\}_{i=1}^{|\mathcal{D}_1|}$; Pre-trained T5 model $\theta$; Pre-training epoch $e_1$; Fine-tuning epoch $e_2$;\\
\textbf{Output}: Trained model $\theta$
\begin{algorithmic}[1] %[1] enables line numbers
\STATE \# Multi-task pre-training
\FOR{\emph{epoch}=1,2,3,..., $e_1$}
\FOR{\emph{Batch $\mathcal{B}$} in \emph{Shuffle($\mathcal{D}_1$)}}
\STATE Optimize model $\theta$ with $\mathcal{B}$=$\{(Z_T, X, y)_i\}_{i=1}^{|\mathcal{B}|}$ in $\mathcal{L}_\theta$
\ENDFOR
\ENDFOR
\STATE 
\STATE \# Single-task fine-tuning
\STATE Perturb $\mathcal{D}_{SG}$ with probabilities $\alpha$ and $\beta$
\FOR{\emph{epoch}=1,2,3,..., $e_2$}
\FOR{\emph{Batch $\mathcal{B}$} in \emph{Shuffle($\mathcal{D}_{SG}$)}}
\STATE Optimize model $\theta$ with $\mathcal{B}$=$\{(Z_T, X, y)_i\}_{i=1}^{|\mathcal{B}|}$ in $\mathcal{L}_\theta$
\ENDFOR
\ENDFOR
\end{algorithmic}
\end{algorithm}

\subsection{Task Formulation}
Given the current user utterance $u_t$, the interaction history $\boldsymbol{h_t}=[u_1, u_2, ..., u_{t-1}]$, the schema of the target database $D=<T,C>$, our goal is to predict the current SQL query $s_t$. Specifically, $D$ consists of a set of tables $T=\{t_i\}_{i=1}^{|T|}$ and columns $C=\{C_i\}_{i=1}^{|T|}$, where for table $t_i$, its corresponding columns $C_i$ is donated as $C_i=\{c_{ij}\}_{j=1}^{|C_i|}$.

\subsection{Multi-task Pre-training}
\subsubsection{Sample Construction}
In this stage, in addition to SG, we propose three auxiliary sub-tasks to improve the main task performance. As shown in Figure \ref{fig:1}, motivated by the T5 training strategy, a training sample input for a specific target task is appended to a task-specific prompt. It is defined as:

\begin{equation}
     \mathcal{S} = (Z_T, X, y)
\label{eq1}
\end{equation}

where $T\in \{SG,RSP,TWP,FUP\}$ stands for the task to which the sample belongs. The task definition is specified in the next section. $Z_T$ denotes the natural language prompt that employs a guidance sentence to indicate each task as shown in Figure \ref{fig:1}.

% \begin{table}[]
% \centering
% \begin{tabular}{l|l}
% \hline
% \textbf{Task}  & \textbf{Prompt}  \\ \hline
% SG              & "translate dialogue to system query:"\\
% RSP             & "translate dialogue to relevant column:"            \\
% FUP            & "translate dialogue to final utterance:"             \\
% TWP & "translate dialogue to turn switch:"            \\ \hline
% \end{tabular}
% \caption{Prompts for specific tasks.}
% \label{table:1}
% \end{table}

$X$ is the model input which is the concatenation of the current user utterance, previous user utterances, and SQL queries, as well as the schema. It is arranged as:
\begin{equation}
     X = [u_1, s_1, u_2, s_2,...,u_t, D]
\label{eq2}
\end{equation}

The schema $D$ is formatted as follows:
\begin{equation}
     D = [t_1, c_{11}, c_{12},...,t_2,c_{21},c_{22},...,... ]
\label{eq3}
\end{equation}

And $y$ is the model output text for the task $T$, which is simplified in Figure \ref{fig:1}.

\begin{table*}[]
\centering
\renewcommand\arraystretch{1.1}
\resizebox{\textwidth}{!}{
\setlength{\tabcolsep}{4mm}{
\begin{tabular}{ccccccccccc}
\hline
\multirow{2}*{\textbf{Dataset}} & \multicolumn{2}{c}{\textbf{Interaction Num}} & \multicolumn{2}{c}{\textbf{Question-SQL Num}}  & \multicolumn{2}{c}{\textbf{RSP Num}} & \multicolumn{2}{c}{\textbf{FUP Num}} & \multicolumn{2}{c}{\textbf{TWP Num}} \\
\cline{2-11}
& Train & Test & Train & Test & Train & Test & Train & Test & Train & Test  \\ \hline
Spider &  / & / & 7000 & 1034 & 7000 & 1034 & / & / & / & / \\ \hline
SparC & 3034 & 422 & 9025 & 1203 & 9025 & 1203 & 2960 & 410 & 5871 & 765 \\ \hline
CoSQL & 2164 & 292 & 7485 & 1008 & 7485 & 1008 & 469 & 69 & 5092 & 704 \\ \hline
\end{tabular}
}}
\caption{Detailed data statistics on training and evaluation. Since Spider is a single-turn dataset, experimental data involving interactions are treated as ignored for this dataset (marked with "/").}
\label{table:1}
\end{table*}

\subsubsection{Training Objective}
The model is trained with a unified maximum likelihood objective for each training sample. Given a training sample $\mathcal{S} = (Z_T, X, y)$, its loss is calculated as: 
\begin{equation}
     \mathcal{L}_\theta = -\sum_{l=1}^LlogP_\theta(y_l|y_{<l};Z_T,X) 
\label{eq4}
\end{equation}

where $L$ denotes the length of the target sequence and $\theta$ is the model parameters.

\subsubsection{Training Tasks}
The pre-training stage focuses on four tasks: the main task SG as well as the three auxiliary tasks RSP, TWP, and FUP. 
\begin{itemize}
    \item \textbf{RSP} is defined as generating the tables and columns that appeared in the current SQL query. This sub-task only needs to consider the to-be-appeared columns and tables without taking the generated SQL structure into account, which can effectively prevent the model from confusing some similar tables and columns.
    \item \textbf{TWP} is introduced to enhance the model by modeling the conversation flow between the adjacent SQL query pair. This task is to predict what modification is made to the current SQL compared with its last SQL. Notably, TWP is originally modeled as a binary classification for each type while it is modeled as a generation task in this paper. If there are multiple modification operations, each operation is separated by the special token $\mathcal{<}/s\mathcal{>}$.
    \item \textbf{FUP} is proposed to generate the final utterance for the whole conversation. The final utterance comes from the original dataset that aims to summarize the information of the whole conversation, so it could be regarded as CQR for the last-turn user utterance in this paper.
\end{itemize}

\subsection{Single-task Fine-tuning}
After the pre-training stage, we further fine-tune the model with the task-specific labeled data with the same learning objective Eq.\ref{eq4}.

\subsubsection{SQL Perturbation}
Since we consider the SQL information from previous turns when generating the current SQL, the model would be hampered by the error propagation problem. Therefore,  we introduce SQL perturbation to reduce the over-dependency of the model on previously generated SQL queries. Specifically, we would replace the previous ground truth SQLs with its corresponding perturbed SQL in the input with a probability $\alpha$. When constructing the perturbed SQL, we randomly sample $\beta$\ tokens of the ground truth SQL and conduct one of the following perturbations:
\begin{itemize}
    \item Randomly modify or add columns of the same table in SELECT sub-clause.
    \item Randomly modify the table in the JOIN structure (i.e., swap “t1” and “t2”).
    \item Randomly modify the “*” column to other columns.
    \item Swap “asc” and “desc”.
\end{itemize}

\subsection{Inference}
During the inference process, we utilize the fine-tuned model to predict the SQL queries for each conversation. When predicting the current SQL, different from the training step where the previous ground truth SQLs or perturbed SQLs are concatenated, we only concatenate the previous predicted SQLs to simulate real application scenarios.

\section{Experiments} 
\subsection{Datasets} 
\subsubsection{Pre-training Datasets} We use context-independent dataset Spider \cite{DBLP:conf/emnlp/YuZYYWLMLYRZR18}, conversational datasets SparC \cite{DBLP:conf/acl/YuZYTLLELPCJDPS19}, CoSQL \cite{DBLP:conf/emnlp/YuZELXPLTSLJYSC19} for the pre-training stage. We create intermediate-labeled data for the proposed three auxiliary tasks on the three datasets above. Specifically, we only perform RSP and SG tasks for Spider because it is context-independent while conducting all three tasks for SparC and CoSQL. Detailed data statistics are shown in Table \ref{table:1}.
\subsubsection{Evaluation Datasets} Since we focus on the conversational text-to-SQL task, SparC and CoSQL datasets are utilized to evaluate the proposed MIGA performance.

\subsection{Experimental Setups}
\subsubsection{Evaluation Metrics} Following previous works \cite{DBLP:journals/corr/abs-2201-05966, DBLP:journals/corr/abs-2205-06983, DBLP:journals/corr/abs-2205-07686}, Interaction match (IM), which requires all output SQL queries in interaction to be valid, is the primary metric used to assess model performance. Every single question's accuracy is also assessed using Question Match (QM).

\subsubsection{Implementation} We implement MIGA based on HuggingFace’s Transformers\footnote{https://github.com/huggingface/transformers} with PyTorch\footnote{https://pytorch.org/} in 2 NVIDIA A100 GPUs. The models are initialized with the pre-trained T5 model of 3B size\footnote{https://huggingface.co/t5-3B}. Epochs and learning rates for 2 training steps are set as (15, 50) and (1e-4, 5e-5). Batch size and optimizer are set as 64, and AdaFactor \cite{DBLP:conf/icml/ShazeerS18} respectively. Following BERT \cite{kenton2019bert}, $\beta$ in SQL perturbation is 15\%. As for the other probability $\alpha$, we set it as 0.15 in this paper after searching.

\subsection{Main Results}

\begin{table}[]
\centering
\renewcommand\arraystretch{1.1}
\setlength{\tabcolsep}{2.3mm}{
\begin{tabular}{lll}
\hline
\textbf{Method} & \textbf{QM(\%)} & \textbf{IM(\%)} \\ \hline
CD-seq2seq \cite{DBLP:conf/acl/YuZYTLLELPCJDPS19} & 21.9 & 8.1 \\
EditSQL \cite{DBLP:conf/emnlp/ZhangYESXLSXSR19} & 47.2 & 29.5 \\
IST-SQL \cite{DBLP:conf/aaai/WangLZ021} & 47.6 & 29.9 \\
GAZA \cite{zhong2020grounded} & 48.9 & 29.7 \\
IGSQL \cite{cai2020igsql} & 50.7 & 32.5 \\
RichContext \cite{liu2021far} & 52.6 & 29.9 \\
TreeSQL V2 \cite{wang2021interactive} & 52.6 & 34.4 \\
R$^2$SQL \cite{hui2021dynamic} & 54.1 & 35.2 \\
DELTA \cite{DBLP:conf/acl/ChenCLCMWY21} & 58.6 & 35.6 \\
SCORE \cite{DBLP:conf/iclr/0009ZPMA21} & 62.2 & 42.5 \\
HIE-SQL \cite{DBLP:conf/acl/ZhengWDWL22} & 64.7 & 45.0 \\
RAT-SQL+TC \cite{DBLP:journals/corr/abs-2112-08735} & 64.1 & 44.1 \\
CQR-SQL \cite{DBLP:journals/corr/abs-2205-07686} & \textbf{67.8} & 48.1 \\ \hline
UNIFIEDSKG \cite{DBLP:journals/corr/abs-2201-05966} & 61.5 & 41.9 \\
RASAT \cite{DBLP:journals/corr/abs-2205-06983} & 64.2 & 43.8 \\
RASAT-PICARD \cite{DBLP:journals/corr/abs-2205-06983} & 66.7 & 47.2 \\ \hline
MIGA(Ours) & 65.4 & 48.6 \\
MIGA+PICARD(Ours) & 67.3 & \textbf{48.9}  \\ \hline
\end{tabular}
}
\caption{Main results on the SparC dev set. The methods in the middle box are all T5-3B-based.}
\label{table:2}
\end{table}

Table \ref{table:2} and Table \ref{table:3} report the conversational text-to-SQL results of different models on SparC and CoSQL datasets. Our proposed MIGA achieves the best IM accuracy on two datasets, obtaining 48.9\% and 29.8\% IM accuracy on the development sets of SparC and CoSQL. The SOTA IM scores indicate the effectiveness of MIGA for the conversational text-to-SQL task. Specifically, our model is based on a single T5-based model architecture that does not require a well-designed relation-aware encoder or tree-based decoder, as well as additional human-annotated data. The intermediate-labeled data used in this paper is self-constructed and derived from the original training samples. This demonstrates the benefits of our model.

Compared with T5-based methods (the middle boxes in Table \ref{table:2} and Table \ref{table:3}), MIGA could obtain significant and consistent improvements in IM scores ranging from 1.7\% to 7.0\% and QM scores ranging from 0.6\% to 5.8\%. This indicates that our training method can stimulate the potential of the T5 pre-trained models for the target task to a greater extent. Besides, PICARD can well constrain the decoding steps of T5 models which can further boost the model performance. When augmented with PICARD, MIGA+PICARD brings a significant improvement on QM metric to the original MIGA. For a fair comparison, we compare MIGA against the version of RASAT \emph{w/o} PICARD as reported in the original paper. The results indicate that MIGA significantly outperforms RASAT, especially in IM accuracy of 4.8\% without PICARD. 

\begin{table}[]
\centering
\renewcommand\arraystretch{1.1}
\setlength{\tabcolsep}{2.3mm}{
\begin{tabular}{lll}
\hline
\textbf{Method} & \textbf{QM(\%)} & \textbf{IM(\%)} \\ \hline
EditSQL \cite{DBLP:conf/emnlp/ZhangYESXLSXSR19} & 39.9 & 12.3 \\
GAZA \cite{zhong2020grounded} & 42.0 & 12.3 \\
IGSQL \cite{cai2020igsql} & 44.1 & 15.8 \\
RichContext \cite{liu2021far} & 41.0 & 14.0 \\
IST-SQL \cite{DBLP:conf/aaai/WangLZ021}  & 44.4 & 14.7 \\
R$^2$SQL \cite{hui2021dynamic} & 45.7 & 19.5 \\
DELTA \cite{DBLP:conf/acl/ChenCLCMWY21} & 51.7 & 21.5 \\
SCORE \cite{DBLP:conf/iclr/0009ZPMA21} & 52.1 & 22.0 \\
HIE-SQL \cite{DBLP:conf/acl/ZhengWDWL22}  & 56.4 & 28.7 \\
CQR-SQL \cite{DBLP:journals/corr/abs-2205-07686} & 58.4 & 29.4 \\ \hline
T5-3B (cholak et al. 2021) & 53.8 & 21.8 \\
PICARD (cholak et al. 2021) & 56.9 & 24.2 \\
UNIFIEDSKG \cite{DBLP:journals/corr/abs-2201-05966} & 54.1 & 22.8 \\
RASAT \cite{DBLP:journals/corr/abs-2205-06983} & 56.2 & 24.9 \\
RASAT-PICARD \cite{DBLP:journals/corr/abs-2205-06983} & 58.8 & 26.3 \\ \hline
MIGA(Ours) & 57.9 & 29.6 \\ 
MIGA+PICARD(Ours) & \textbf{59.0} & \textbf{29.8} \\ \hline
\end{tabular}
}
\caption{Main results on the CoSQL dev set. The methods in the middle box are all T5-3B-based.}
\label{table:3}
\end{table}

To further verify the SQL prediction effectiveness of the proposed MIGA, we compare the performance of some baseline methods with MIGA on different interaction turns and SQL difficulties. As shown in Tables \ref{table:4} and \ref{table:5}, the accuracy of SQL prediction decreases as the interaction turn and difficulty increases due to the long-range dependency and complex SQL structures. MIGA achieves competitive performance on different turns and difficulties, outperforming most of the baseline models except CQR-SQL. Typically, Since CQR-SQL focuses on CQR, it can effectively alleviate the dependency of information extraction on the conversation and previous SQLs, thus achieving superior performance on QM metrics especially for later turns (Turn 3 and Turn 4) and high difficulties (Hard and Extra). This also indicates the importance of CQR for the target task. Unfortunately, as the original CQR data in CQR-SQL is a newly introduced dataset and it is not publicly released, it is impossible to include this data in our MIGA model. Moreover, the data size of the proposed FUP task in this paper is only 3908, and the data-hungry problem limits the ability of the CQR task to boost MIGA. How to extend the data size of the FUP task is left for future work.

\subsection{Analysis}
\subsubsection{Ablation Study}
We conduct experiments on different variants of the proposed framework to investigate the contributions of different components. There are 5 different variants: 1) \emph{w/o} RSP: we remove the RSP task in the multi-task pre-training stage; 2) \emph{w/o} TWP: we remove the TWP task in the multi-task pre-training stage; 3) \emph{w/o} FUP: we remove the FUP task in the multi-task pre-training stage; 4) \emph{w/o} SQL perturbation: we remove SQL perturbation in the fine-tuning step and fine-tune the model as in the pre-training stage; and 5) \emph{w/o} Spider/CoSQL: we remove the training samples of the corresponding dataset in the pre-training stage. All the variants are evaluated on the SparC dev set.

\begin{table}[]
\centering
\renewcommand\arraystretch{1.1}
\resizebox{\columnwidth}{!}{
\begin{tabular}{lcccc}
\hline
\textbf{Method} & \textbf{Turn 1} & \textbf{Turn 2} & \textbf{Turn 3} & \textbf{Turn 4+} \\ \hline
EditSQL &	62.2 &	40.1 &	36.1 &	19.3 \\
IG-SQL &	63.2 &	50.8 &	39.0 &	26.1 \\
R$^2$SQL &	67.7 &	55.3 &	45.7 &	33.0 \\
RAT-SQL+TC &	75.4 &	64.0 &	54.4 &	40.9 \\
CQR-SQL &	75.6 &	68.7 &	58.9 &	53.9 \\ \hline
MIGA+PICARD &	74.9 &	68.5 &	58.9 &	52.8 \\
\hline
\end{tabular}
}
\caption{Detailed QM accuracy performance on different interaction turns of SparC dev set.}
\label{table:4}
\end{table}

\begin{table}[]
\centering
\renewcommand\arraystretch{1.1}
\resizebox{\linewidth}{!}{
\begin{tabular}{lcccc}
\hline
\textbf{Method} & \textbf{Easy} & \textbf{Medium} & \textbf{Hard} & \textbf{Extra} \\ \hline
EditSQL &	68.8 &	40.6 & 26.9 &	12.8 \\
IG-SQL &	70.9 &	45.4 &	29.0 &	18.8 \\
R$^2$SQL &	75.5 &	51.5 &	35.2 &	21.8 \\
CQR-SQL &	80.7 &	68.3 &	46.2 &	43.3 \\ \hline
MIGA+PICARD &	81.8 &	66.7 &	44.8 &	41.8 \\
\hline
\end{tabular}
}
\caption{Detailed QM accuracy performance on different SQL difficulties of the SparC dev set.}
\label{table:5}
\end{table}

Table \ref{table:7} shows the results of the variants and MIGA. We can see that generally, all components contribute to the target task. The full model consistently achieves the best performance on the SparC dev set. Some in-depth analyses are shown as follows.
\begin{itemize}
    \item Removing RSP task leads a performance drops of 4.0\% and 4.4\% in QM and IM metrics respectively. This task can effectively model the information of the tables and columns that appeared in predicting SQL, which further avoids the confusion of similar columns and tables.
    \item If removing the TWP task, two metrics on SparC drop by 3.4\% and 3.2\%. TWP can obtain information about the interaction between the questions and SQLs, thus reducing the difficulty of the current SQL prediction. However, this task models the modifications of the previously generated SQL which may cause error propagation in the inferences stage to some extent.
    \item Adding FUP task can get a performance gain of 2.1\% and 1.8\% on two metrics. However, the training sample size for this task is relatively small, resulting in a relatively small performance improvement. Constructing more training samples for this task may further improve the performance of the model.
    \item SQL perturbation can significantly alleviate the error propagation problem, which achieves performance gains of (1.0\%, 1.0\%) on two metrics.
    \item Removing either of the pre-trained datasets leads to a performance drop. Specifically, the CoSQL dataset seems to contribute more than the Spider dataset on IM metric, for Spider is a context-independent dataset which could only boost the first-turn SQL prediction potentially.
\end{itemize}

\subsubsection{Model Size Impact}
\begin{figure}
  \centering
  \includegraphics[width=0.48\textwidth]{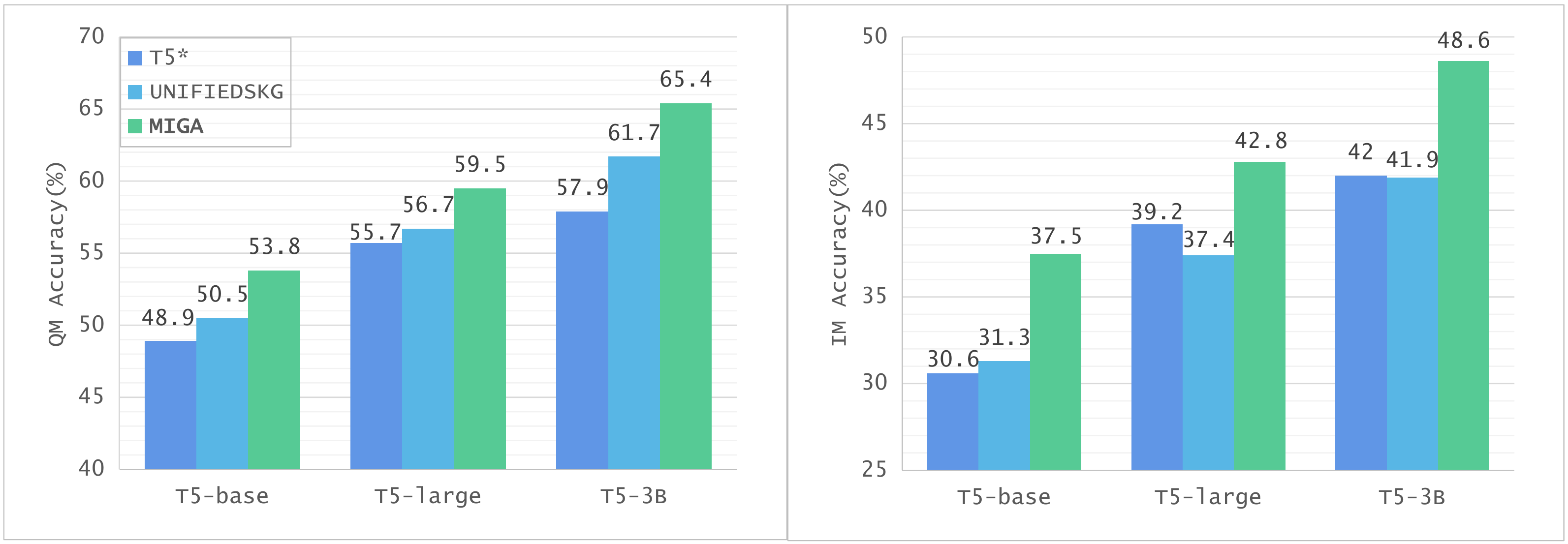} 
  \caption{Results on the SparC dev set for different T5 pre-trained model sizes. * indicates our re-implementation. The results of UNIFIEDSKG come from the original paper.} 
  \label{fig:2} 
\end{figure}

To evaluate the proposed framework on pre-trained models with different sizes, we implement MIGA and fine-tune the vanilla T5 pre-trained models on 3 different model sizes (T5-base, T5-large, T5-3B) and evaluate them on the SparC dev set. The results are shown in Figure \ref{fig:2}. Commonly, when the model size increases, the model performance also increases significantly, demonstrating the powerful semantic modeling ability of large PLMs. In addition, the base-size MIGA model outperforms the large-size vanilla fine-tuned T5 model, and the same situation occurs for the large-size MIGA, which again illustrates the effectiveness of MIGA. The three proposed auxiliary tasks effectively stimulate the potential of the T5 pre-trained models and add more training data. This can encourage the pre-trained model to better adapt to the target task. 

\begin{table}[]
\centering
\renewcommand\arraystretch{1.1}
\setlength{\tabcolsep}{4mm}{
\begin{tabular}{lcc}
\hline
\textbf{Model} & \textbf{QM} & \textbf{IM} \\ \hline
MIGA & 65.4  & 48.6 \\ \hline
\emph{w/o} RSP & 61.4 (-4.0) & 44.2 (-4.4) \\
\emph{w/o} TWP & 62.0 (-3.4) & 45.4 (-3.2) \\ 
\emph{w/o} FUP & 63.3 (-2.1) & 46.8 (-1.8) \\
\emph{w/o} SQL perturbation & 64.4 (-1.0) & 47.6 (-1.0) \\ \hline
\emph{w/o} Spider & 63.8 (-1.6) & 47.3 (-1.3) \\
\emph{w/o} CoSQL & 63.8 (-1.6) & 45.8 (-2.8) \\ \hline
\end{tabular}
}
\caption{Ablation study.}
\label{table:7}
\end{table}

\begin{figure*}
  \centering
  \includegraphics[width=1\textwidth]{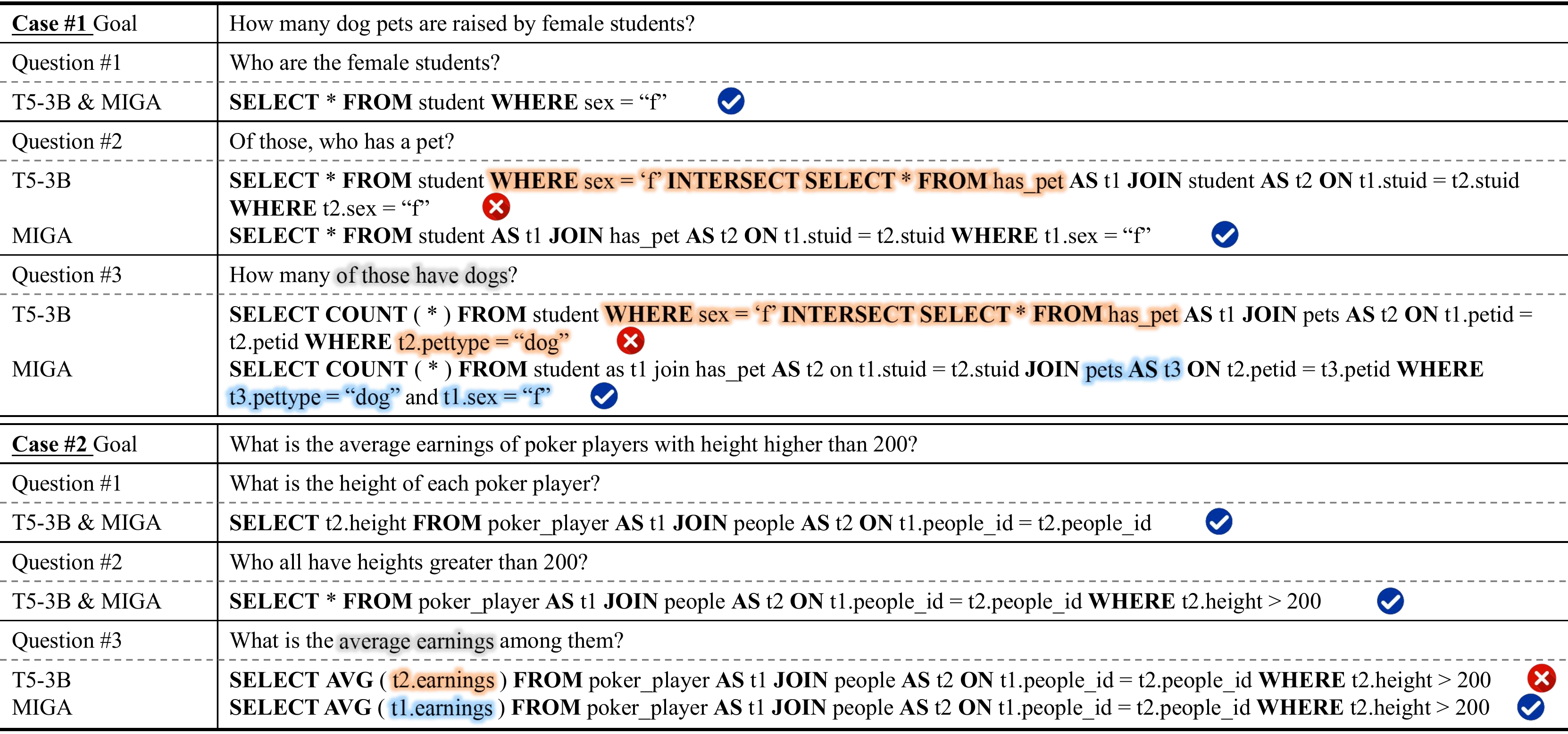} 
  \caption{Case studies on the SparC dev set. MIGA gives correct predictions in two cases while the vanilla T5-3B model fails.} 
  \label{fig:4} 
\end{figure*}

\subsubsection{Case Study}
Figure \ref{fig:4} presents two cases derived from the SparC dev set and the corresponding predicted SQL queries of the vanilla fine-tuned T5-3B and MIGA. For \emph{Case \#1}, in \emph{Question \#2}, T5-3B  fails to generate valid SQL for relatively complex JOIN  structures (two-table join), which in turn leads to the failure of more complex JOIN structures (three-table join) in \emph{Question \#3}. While MIGA accurately predicts the JOIN structure and maintains the previous condition {\tt t1.sex="f"} based on the extracted turn switch and related column information. For \emph{Case \#2}, T5-3B tends to confuse the columns of different tables and misidentifies {\tt earnings} as a column of table {\tt people}, while MIGA can correctly identify the column belongs to table {\tt poker\_player} and link it to {\tt t1} by modeling the table and column information.

\subsubsection{Searching of Hyper-parameter $\alpha$}
\begin{figure}[]
  \centering
  \includegraphics[width=0.48\textwidth]{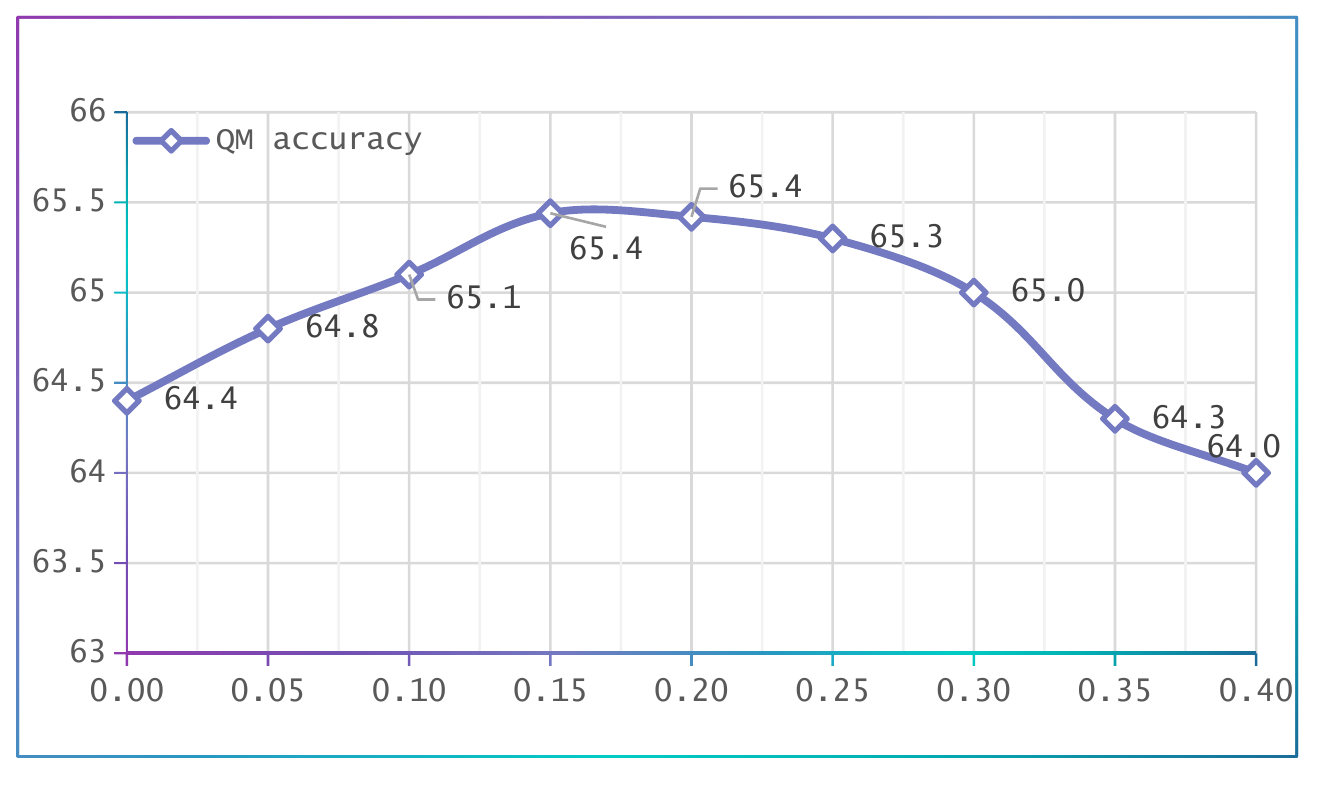} 
  \caption{Results on the SparC dev set of different $\alpha$ values. The experiments are conducted on MIGA with T5-3B size.} 
  \label{fig:5} 
\end{figure}

When no SQL perturbation is used, the QM accuracy of MIGA with 3B size on SparC dev set is 64.4\%. To simulate the real error scenario, we set the search range as $[0.0-0.4]$ and use QM accuracy as the metric for parameter selection. As shown in Figure \ref{fig:5}, when the $\alpha$ value is 0.15, the QM score of the model reaches the peak. In addition, when it exceeds 0.35, the model performance starts to be inferior to the model performance without SQL perturbation.

\subsection{Limitations and Future Directions}
One main concern is that the QM accuracy of our model is potentially inferior to some SOTA methods. One possible reason is that concatenating previously predicted SQL into the current input may cause error propagation. This results in the model prediction somewhat being dependent on the first-turn predicted SQL, i.e., if the first-turn SQL is predicted correctly, the later-turn SQLs tend to be predicted correctly, and vice versa. To further explore this issue, we input the correct first-turn SQL to evaluate the predicted SQLs of subsequent turns. We found that if the first-turn SQL is correct, the accuracy of the subsequent predicted SQLs (especially the second turn) is greatly improved, thus effectively improving the IM score as well. This indicates errors in the first several steps have a huge impact on the final results. To resolve this, we can either improve the accuracy of the first-turn SQL prediction or reduce the dependence of the current-turn SQL prediction on the previous SQL prediction. Although our proposed SQL perturbation could alleviate this problem, we plan to further reduce it in the future.

% \begin{table}[]
% \centering
% \renewcommand\arraystretch{1.2}
% {\resizebox{\linewidth}{!}{\begin{tabular}{lllllll}
% \hline
%  & \textbf{Turn 1} & \textbf{Turn 2} & \textbf{Turn 3} & \textbf{Turn 4+} & \textbf{QM (all)} & \textbf{IM}\\ \hline
% Prediction & 73.7 & 67.3 & 55.9 & 46.6 & 65.4 & 48.6\\
% Ground truth & 100.0 & 73.9 & 58.9 & 47.2 & 77.7 & 60.2 \\ \hline
% \end{tabular}}}
% \caption{Performance on SparC dev set between predicted and ground truth first-turn SQL. Prediction and Ground truth indicate using predicted first-turn SQL and ground truth first-turn SQL respectively. }
% \label{table:8}
% \end{table}

\section{Conclusions}
In this paper, we propose MIGA, a two-stage unified multi-task generation framework, for conversational text-to-SQL. Specifically, in the pre-training stage, we decompose the target task into three sub-tasks and unify them into the same sequence-to-sequence (Seq2Seq) paradigm by adding a task-specific natural language prompt into the input text to indicate each task. Later to adapt the model to the target benchmark, we fine-tune it in the task-specific labeled data and introduce four SQL perturbations to alleviate the error propagation problem. Extensive experiments and analyses on two conversational text-to-SQL benchmarks indicate the effectiveness of MIGA. In future work, we would further explore the strategies to tackle the error propagation problem and extend the data size of the FUP task.

\bibliography{aaai23}

\end{document}